\documentclass[11pt]{article}

\usepackage[final]{acl}

\usepackage{times}
\usepackage{latexsym}

\usepackage[T1]{fontenc}

\usepackage{microtype}

\usepackage{inconsolata}

\usepackage{graphicx}

\usepackage[table]{xcolor}
\usepackage{booktabs}
\usepackage{multirow}
\usepackage{graphicx}
\usepackage{amsmath}
\usepackage{amssymb}

\title{Making Every Verified Token Count: \\Adaptive Verification for MoE Speculative Decoding}

\author{
 \textbf{Lehan Pan\textsuperscript{1}},
 \textbf{Ziyang Tao\textsuperscript{1}},
 \textbf{Xiao Wang\textsuperscript{2}},
 \textbf{Zhimeng Zhao\textsuperscript{2}},
 \textbf{Meng Wang\textsuperscript{2}},
 \textbf{Yanyong Zhang\textsuperscript{1}\thanks{The Corresponding Author.}}
\\
\textsuperscript{1}{University of Science and Technology of China}
\\
\textsuperscript{2}{Tianyijiaotong Technology Ltd., Suzhou, China}
\\
{\tt \{lhpan,ziyangtao\}@mail.ustc.edu.cn, yanyongz@ustc.edu.cn}
\\
{\tt \{xiao.wang,zhimeng.zhao,meng.wang\}@tyjt-ai.com}
}

\begin{document}
\maketitle

\begin{abstract}
Tree-based speculative decoding accelerates autoregressive generation by verifying multiple draft candidates in parallel, but this advantage weakens for sparse Mixture-of-Experts (MoE) models. As the draft tree grows, different branches activate different experts, expanding the union of activated experts and substantially increasing target-side verification cost. We propose EVICT, a training-free, hyperparameter-free, and lossless adaptive verification method for MoE speculative decoding. EVICT makes every verified token count by truncating the draft tree before target verification and retaining only the cost-effective prefix. It leverages fine-grained drafter signals to estimate candidate benefit, combines them with offline-profiled verification cost, and remains highly compatible with the high-performance graph-based serving framework SGLang. Extensive experiments on diverse MoE backbones and benchmarks show that EVICT achieves up to 2.35$\times$ speedup over autoregressive decoding and an average 1.21$\times$ speedup over the state-of-the-art baseline EAGLE-3, while significantly reducing unnecessary expert activations during verification.
\end{abstract}
\section{Introduction}

Mixture-of-Experts (MoE) offers an attractive way to scale language models by increasing parameter capacity without proportionally increasing per-token computation through conditional activation of experts \citep{shazeer2017outrageously, lepikhin2020gshard, fedus2022switch, du2022glam, liu2024deepseek, zeng2026glm, team2025kimi, cao2026qwen3}. In parallel, speculative decoding has emerged as a lossless technique for accelerating autoregressive generation \citep{pmlr-v202-leviathan23a, chen2023acceleratinglargelanguagemodel}, and tree-based variants further improve efficiency by organizing drafted candidates into a token tree that can be verified in parallel \citep{miao2024specinfer, li2024eagle, cai2024medusa, chen2024sequoia}. These two lines of progress naturally suggest speculative decoding as a promising acceleration strategy for modern MoE language models.

\begin{figure}[t]
    \centering
    \includegraphics[width=\columnwidth]{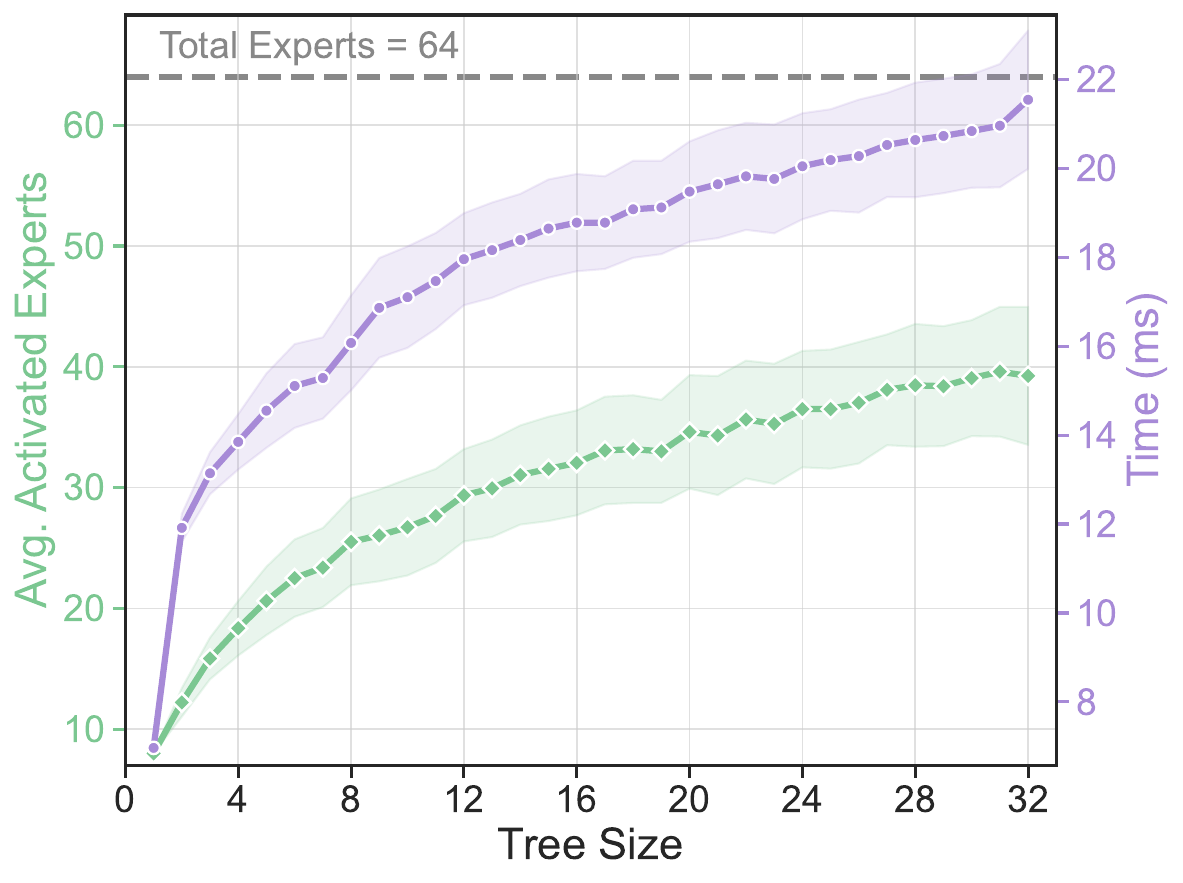}
    \caption{Average activated experts and per-iteration latency versus tree size of Qwen3-30B-A3B, averaged across a full inference run. Larger trees activate more experts, thus incur higher iteration latency.}
    \label{fig:experts_time_growth}
\end{figure}

\begin{figure}[t]
    \centering
    \includegraphics[width=\columnwidth]{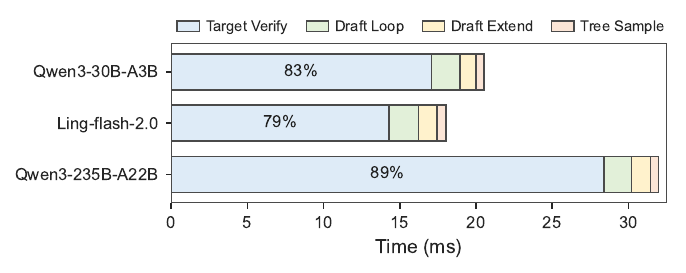}
    \caption{Per-iteration latency breakdown averaged across a full inference run. In all cases, Target Verify dominates the latency budget, highlighting target verification as the main bottleneck.}
    \label{fig:breakdown}
\end{figure}

However, the standard intuition from dense models does not directly carry over to sparse MoE targets \citep{huang2025moesd}. For MoE, each token activates only a small subset of experts, but tree verification aggregates the union of experts activated by all verified nodes \citep{mcdanel2026moe}. As a result, enlarging the draft tree can quickly increase the number of activated experts and the per-iteration latency. As shown in Figure~\ref{fig:experts_time_growth}, larger trees activate more experts and incur higher iteration latency, while Figure~\ref{fig:breakdown} shows that target verification already dominates the latency budget of speculative decoding under moderate tree size. Together, these observations reveal a central tension for MoE speculative decoding: a larger tree may improve draft coverage, yet simultaneously worsen the very bottleneck that determines end-to-end latency.

This problem is especially pronounced in the small-batch, low-latency regime, where experts are sparsely activated. In this setting, the goal should not be to maximize accepted tokens by aggressively enlarging the draft tree. Instead, every draft token should be worth its cost: some draft tokens are valuable because they are likely to contribute to the final accepted path, whereas others add little acceptance benefit but trigger substantial extra expert activation and memory traffic. This perspective differs from prior tree-based speculative decoding methods \citep{li2024eagle2, wang2025opt, liu2026talon}, which mainly optimize draft-tree construction from the acceptance side.

Motivated by this observation, we propose \textbf{EVICT}, a training-free, hyperparameter-free, and lossless \textbf{E}fficient \textbf{V}erification strategy via \textbf{I}dentifying \textbf{C}ost-effective \textbf{T}ree prefixes. EVICT leverages fine-grained drafter signals to estimate the benefit of candidates, combines them with offline-profiled verification cost, and decides at each decoding step how many draft nodes are actually worth verifying. Crucially, EVICT is co-designed with the serving system SGLang \citep{zheng2024sglang}: rather than introducing intricate runtime control that disrupts efficient kernels, it remains compatible with high-performance inference frameworks by pre-capturing verification graphs and fusing the added online logic into the draft graph. Our contributions are threefold: \begin{itemize}
    \item We formulate MoE speculative decoding as a cost-effective verification problem, showing that the key is not to maximize draft acceptance alone, but to select the verification prefix with the best benefit-to-cost trade-off.

    \item We propose EVICT, a MoE-cost-aware adaptive verification method that exploits fine-grained drafter information while remaining compatible with modern serving frameworks.

    \item Across various MoE models and benchmarks, EVICT achieves an average 1.21$\times$ speedup over the state-of-the-art method EAGLE-3 \citep{eagle3}, while substantially reducing unnecessary expert activations during verification. Additional results validate the method across batch sizes, drafting backbones, and a dense target model.
\end{itemize}

\section{Preliminaries}
\label{sec:preliminaries}

We briefly review the tree-based speculative decoding framework used in EAGLE-3 \citep{eagle3}. Its sampling-time verification process forms the theoretical foundation of our method, and we further analyze why MoE models can suffer performance degradation under speculative decoding.

\subsection{Tree-Based Speculative Decoding}\label{sec:tree_sd}

\begin{figure}[t]
    \centering
    \includegraphics[width=\columnwidth]{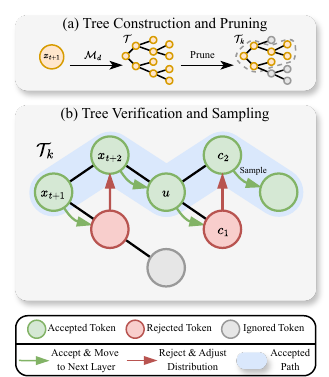}
    \caption{Overview of tree-based speculative decoding, including draft-tree construction, target verification, and tree-based sampling.}
    \label{fig:tree_and_path}
\end{figure}

\begin{figure*}[t]
    \centering
    \includegraphics[width=\textwidth]{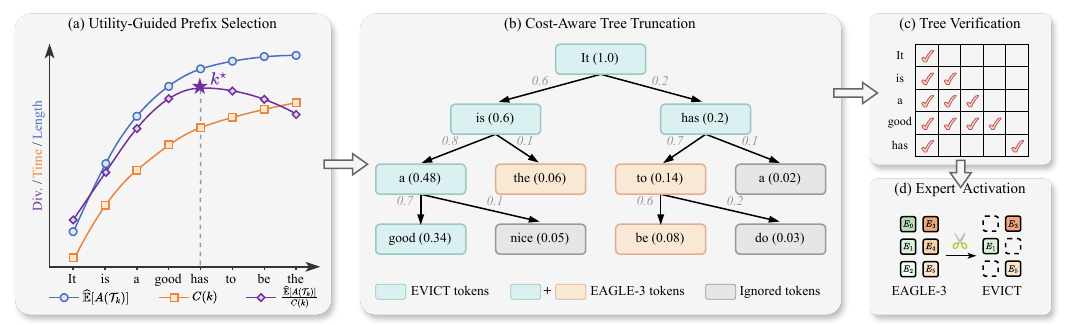}
    \caption{(a) EVICT selects the verification budget by balancing the estimated accepted length against the profiled verification cost, and chooses the utility-maximizing prefix size. (b) Given the selected budget, EVICT truncates the draft tree into an ancestor-closed cost-effective prefix, while EAGLE-3 verifies additional low-utility nodes under a fixed tree budget. (c) The retained prefix is verified in a single target-model forward pass using a tree attention mask. (d) By verifying fewer draft nodes, EVICT reduces the union of activated experts in the MoE target model, thereby lowering target-side verification cost.}
    \label{fig:method}
\end{figure*}

Given a target model $\mathcal{M}_t$ and a draft model $\mathcal{M}_d$, at time $t$, one speculative decoding step starts from a prefix $x_{\le t}$. The target model first produces the next-token distribution
\begin{equation}
    p_{t+1}(\cdot)=P_{\mathcal{M}_t}(\cdot \mid x_{\le t}),
\label{eq:next_tok_dist}
\end{equation}
from which one token $x_{t+1}$ is sampled. Conditioned on $x_{\le t+1}$, $\mathcal{M}_d$ repeatedly extends a fixed number of draft tokens to produce a well-structured token tree $\mathcal{T}$ rooted at $x_{t+1}$, with the same number of tokens at each layer. The tree is then pruned, as described in Section~\ref{sec:prune} and Fig.~\ref{fig:tree_and_path}~(a), to meet a predefined token budget $k$, yielding $\mathcal{T}_k$. The target model then verifies all nodes in $\mathcal{T}_k$ in a single forward pass, using the attention mask shown in Fig.~\ref{fig:method}~(c). For each node $v \in \mathcal{T}$ and $v^{\prime} \in \mathcal{T}_k$, we obtain the draft and target distributions
\begin{equation}
    \begin{gathered}
        q_{t+1}^{\mathcal{T}}(v) = P_{\mathcal{M}_d}(v \mid x_{<v}), \\
        p_{t+1}^{\mathcal{T}_k}(v^{\prime}) = P_{\mathcal{M}_t}(v^{\prime} \mid x_{<v^{\prime}}),
    \end{gathered}
\end{equation}
where $x_{<v}$ and $x_{<v^{\prime}}$ denote the prefix tokens of $v$ and $v^{\prime}$, respectively.

Under sampling, the verified tree is consumed layer by layer. Once verification reaches an intermediate layer, the shallower part of the tree has already collapsed into a single surviving path ending at intermediate node $u$, as illustrated in Fig.~\ref{fig:tree_and_path} (b). The local decision therefore only concerns its children $\operatorname{Child}(u)$ sequentially. For each $c \in \operatorname{Child}(u)$, the verifier accepts $c$ with probability $p_{t+1}^{\mathcal{T}}(c)$. Upon acceptance, the surviving path is extended to $c$, and verification advances to the next layer. Upon rejection, the mass of $c$ is removed, updating the distribution for each $w \in \operatorname{Child}(u)$:
\begin{equation}
    p_{t+1}^{\mathcal{T}}(w)=
    \begin{cases}
        0, & w=c,\\[2mm]
        \dfrac{p_{t+1}^{\mathcal{T}}(w)}{1-p_{t+1}^{\mathcal{T}}(c)}, & w\neq c.
    \end{cases}
\end{equation}
The verifier then continues with the next sibling of $c$ using the updated distribution. If all siblings are rejected, a fresh token is sampled from the final residual distribution. Overall, this sampling logic preserves the original target distribution.

After the draft tree yields one accepted path, decoding returns to Equation~\eqref{eq:next_tok_dist} for the next speculative iteration.

\subsection{MoE Verification Cost}

We consider the case where the target model $\mathcal{M}_t$ is a sparsely activated MoE model with $N$ experts. At each MoE layer, given a hidden state $\mathbf{h} \in \mathbb{R}^d$, the router computes expert scores $\mathbf{W}_g \mathbf{h} \in \mathbb{R}^N$, where $\mathbf{W}_g \in \mathbb{R}^{N \times d}$ and each row of $\mathbf{W}_g$ can be viewed as the centroid vector of the corresponding expert. The router then selects the top-$k$ experts:
\begin{equation}
    \mathcal{E}(\mathbf{h})=\operatorname{TopK}(\mathbf{W}_g \mathbf{h}, k),
\end{equation}
where $\operatorname{TopK}(\cdot, k)$ returns the indices of the $k$ largest values. The selected scores are normalized into expert weights, e.g., by $\mathrm{softmax}$ or $\operatorname{sigmoid}$ depending on the architecture, and the layer output is the corresponding weighted sum. Thus, each token activates only $k \ll N$ experts at an MoE layer.

In speculative decoding, however, the target model verifies the entire draft tree $\mathcal{T}$ in one pass. Let $\mathcal{H}=\{\mathbf{h}_{v}\mid v \in \mathcal{T}\}$ be the hidden states of all verified tree nodes. The activated experts at this layer are then
\begin{equation}
    \mathcal{E}(\mathcal{H})=\bigcup_{\mathbf{h} \in \mathcal{H}} \mathcal{E}(\mathbf{h}).
\end{equation}
As the tree grows, this union can expand rapidly because different branches may route to different experts. Consequently, speculative decoding can substantially increase the number of activated experts in a target pass, further aggravating the memory-bandwidth bottleneck of LLM inference \citep{yuan2024llm} and offsetting its speedup. 

The efficiency of MoE speculative decoding varies across expert-unsaturated, expert-saturated, and compute-bound regimes. EVICT primarily targets the expert-unsaturated, memory-bound regime; Appendix~\ref{app:regime_analysis} provides the detailed roofline analysis.

\section{Method}
\label{sec:method}

\subsection{Accepted Length Estimation}
\label{sec:accepted_length_estimation}

The key quantity for tree selection is the expected number of tokens committed in one speculative step. Let $A(\mathcal{T})$ denote the accepted length of a draft tree $\mathcal{T}$. Under the standard tree-based speculative decoding rule in Section~\ref{sec:tree_sd}, the expected accepted length can be written as
\begin{equation}
    \mathbb{E}[A(\mathcal{T})]
    =
    \sum_{v \in \mathcal{T}}
    \prod_{u \in \operatorname{Path}(x_{t+1}, v)}
    p_{t+1}^{\mathcal{T}}(u),
    \label{eq:exact_accept_len}
\end{equation}
where $\operatorname{Path}(x_{t+1}, v)$ denotes the path from the root $x_{t+1}$ to node $v$. We provide the derivation in Appendix~\ref{app:accepted_length_proof}.

Equation~\ref{eq:exact_accept_len} is exact but requires target probabilities after verification. To estimate this quantity before verification, we replace each target factor with the corresponding draft probabilities and define the score of a node as
\begin{equation}
    \operatorname{Score}(v)
    :=
    \prod_{u \in \operatorname{Path}(x_{t+1}, v)}
    q_{t+1}^{\mathcal{T}}(u).
    \label{eq:node_score}
\end{equation}
Using draft probabilities as empirical proxies for target-side acceptance, a pre-verification estimate of tree quality for adaptive truncation is given by
\begin{equation}
    \widehat{\mathbb{E}}[A(\mathcal{T})]
    =
    \sum_{v \in \mathcal{T}} \operatorname{Score}(v).
    \label{eq:estimated_accept_len}
\end{equation}

Appendix~\ref{app:estimator_validation} validates this approximation across low- and high-acceptance workloads using calibration analysis.

\subsection{Utility-Guided Tree Truncation}

To balance the gain in expected accepted length against the growing verification cost, we first prune the well-structured draft tree under a fixed budget (Section~\ref{sec:prune}), then select the best verification prefix via a utility objective (Section~\ref{sec:choose_prefix}), and finally enable efficient online decisions through offline cost profiling (Section~\ref{sec:prof_cost}).

\subsubsection{Pruning a Well-Structured Draft Tree}\label{sec:prune}

Existing tree-based speculative decoding methods, such as EAGLE-3 \citep{eagle3} and OPT-Tree \citep{wang2025opt}, prune the well-structured draft tree $\mathcal{T}$ before verification and retain the top-$k$ nodes ranked by cumulative scores, yielding a truncated tree $\mathcal{T}_k$ under a verification budget $k$. This truncation is well defined because the cumulative score of a child cannot exceed that of its parent, making the selected node set automatically ancestor-closed and hence a valid tree. Moreover, since the estimated accepted length is the sum of node scores, the resulting $\widehat{\mathbb{E}}[A(\mathcal{T}_k)]$ is optimal among all valid $k$-node subtrees of $\mathcal{T}$.

\subsubsection{Choosing Best Prefix for Verification}\label{sec:choose_prefix}

However, maximizing $\widehat{\mathbb{E}}[A(\mathcal{T}_k)]$ alone is not sufficient for end-to-end acceleration of an MoE model, because verification cost also grows with $k$. Let $C(k)$ denote the latency of a whole speculative iteration when the top-$k$ draft nodes are kept and verified, and let $C_{\mathrm{AR}}$ denote the per-token latency of standard autoregressive decoding on the same target model. Following prior work \citep{saxena2025utility}, we define the expected speculation utility as
\begin{equation}
    U(k)
    :=
    \frac{C_{\mathrm{AR}}\,\widehat{\mathbb{E}}[A(\mathcal{T}_k)]}{C(k)}
    \label{eq:spec_utility}
\end{equation}
and maximize this iteration-level utility, since it is equivalent to minimizing average time per output token (TPOT). Under this average-case view, $C_{\mathrm{AR}}$ is constant for a fixed target model. We therefore select the best prefix
\begin{equation}
    k^\star
    =
    \arg\max_{1 \le k \le \texttt{draft\_tokens}}
    \frac{\widehat{\mathbb{E}}[A(\mathcal{T}_k)]}{C(k)},
    \label{eq:best_prefix}
\end{equation}
and verify $\mathcal{T}_{k^\star}$ at each iteration.

In difficult contexts, node confidence drops quickly and $\widehat{\mathbb{E}}[A(\mathcal{T}_k)]$ saturates while $C(k)$ grows, leading to early truncation; in easy contexts, high-confidence nodes persist in deeper layers, so $\widehat{\mathbb{E}}[A(\mathcal{T}_k)]$ keeps increasing and can justify a larger $C(k)$. The verification budget thus adapts automatically to the local difficulty of each decoding step.

\begin{table*}[t]
\centering
\tabcolsep=0.6em
\scalebox{0.83}{
\begin{tabular}{l|*{6}{>{\centering\arraybackslash}p{4.55em}}|c}
\toprule
Method & Alpaca & GSM8K & HumanEval & QA & MT-Bench & CNN/DM & Average \\
\midrule
\multicolumn{8}{c}{Temperature = 0} \\
\midrule
Vanilla & 149.75 & 143.58 & 143.71 & 144.62 & 143.10 & 142.02 & 144.46 (1.00$\times$) \\
Lookahead & 94.82 & 99.17 & 99.33 & 93.60 & 101.59 & 80.62 & 94.86 (0.66$\times$) \\
\midrule
EAGLE-3 & 173.01 & 208.74 & 209.32 & 152.79 & 171.08 & 143.12 & 176.34 (1.22$\times$) \\
\midrule
+DDD & 156.67 & 201.38 & 203.56 & 140.22 & 159.16 & 133.18 & 165.70 (1.15$\times$) \\
\textbf{+EVICT} & \textbf{183.82} & \textbf{258.35} & \textbf{265.52} & \textbf{162.54} & \textbf{190.34} & \textbf{156.35} & \textbf{202.82 (1.40$\times$)} \\
\midrule
\multicolumn{8}{c}{Temperature = 1} \\
\midrule
Vanilla & 146.27 & 140.14 & 140.33 & 141.81 & 140.13 & 139.71 & 141.40 (1.00$\times$) \\
Lookahead & 92.45 & 98.10 & 97.38 & 89.69 & 99.00 & 79.76 & 92.73 (0.66$\times$) \\
\midrule
EAGLE-3 & 165.50 & 203.51 & 207.00 & 145.62 & 162.26 & 136.80 & 170.11 (1.20$\times$) \\
\midrule
+DDD & 151.18 & 198.93 & 199.54 & 134.23 & 152.31 & 129.72 & 160.99 (1.14$\times$) \\
\textbf{+EVICT} & \textbf{174.84} & \textbf{249.97} & \textbf{259.22} & \textbf{155.60} & \textbf{180.18} & \textbf{151.19} & \textbf{195.17 (1.38$\times$)} \\
\bottomrule
\end{tabular}
}
\caption{Decoding speed (token/s) of our proposed EVICT and baselines on Qwen3-30B-A3B across various benchmarks and temperatures. Values in parentheses indicate the speedup ratio relative to the Vanilla baseline.}
\label{tab:speed_qwen_30b}
\end{table*}

\subsubsection{Profiling Cost before Runtime}\label{sec:prof_cost}

To evaluate Equation~\ref{eq:best_prefix}, we directly profile the end-to-end speculative-step latency $C(k)$ for all feasible $k \in \{1,\ldots, \texttt{draft\_tokens}\}$ at initialization time and store the measurements in a lookup table. Here, $C(k)$ estimates the expected average cost for verification length $k$ under a fixed runtime configuration; it is not an exact per-step prediction for the particular expert set activated by the current verified tokens. Figure~\ref{fig:experts_time_growth} and Appendix~\ref{app:estimator_validation} validates the stability of this average cost profile across workloads. Therefore, pre-profiling and table lookup provide a reliable estimate during online decoding. At runtime, utility maximization reduces to computing prefix cumulative sums of node scores to obtain $\widehat{\mathbb{E}}[A(\mathcal{T}_k)]$ for all $k$, performing element-wise division by the corresponding costs $C(k)$, and taking an $\arg\max$ over $k$.

\subsection{SGLang Integration}\label{sec:sgl_integrate}

To eliminate potential overhead from framework-level inefficiencies and to provide a more convincing evaluation of our method, we build our implementation on top of SGLang \citep{zheng2024sglang}, a state-of-the-art inference system with support for CUDA-graph-based speculative decoding. Since CUDA graph is crucial for inference speedup, our added runtime operations are designed to remain fully and seamlessly compatible with CUDA graph execution.

\paragraph{Target Graph Capture.}
SGLang captures target and draft CUDA graphs once at initialization and replays them during decoding. Because our verification length is adaptive, a single target-verify graph is insufficient. We therefore pre-capture a set of verify graphs for different verification lengths, and at runtime directly dispatch the one matching the selected length without recapture.

\paragraph{Draft Graph Fusion.}
To further minimize the latency introduced by our method, we fuse the additional runtime operations into the draft graph during initialization. After the draft tree is constructed, scores for all candidate nodes become available. We then execute the runtime operations described in Section~\ref{sec:prof_cost} within the draft graph capture region.

\section{Experiments}

\subsection{Experimental Setup}

\paragraph{Models and Datasets.}
We evaluate EVICT on three MoE backbones with different model scales and routing sparsity, as well as a dense target model; detailed configurations are provided in Appendix~\ref{app:models}. Following standard protocols, we use six representative benchmarks: MT-Bench~\citep{mt-bench} for multi-turn dialogue, Alpaca~\citep{alpaca} for instruction following, GSM8K~\citep{gsm8k} for mathematical reasoning, HumanEval~\citep{humaneval} for code generation, QA~\citep{qa} for question answering, and CNN/DM~\citep{cnndm} for summarization.

\begin{figure*}[t]
    \centering
    \includegraphics[width=\textwidth]{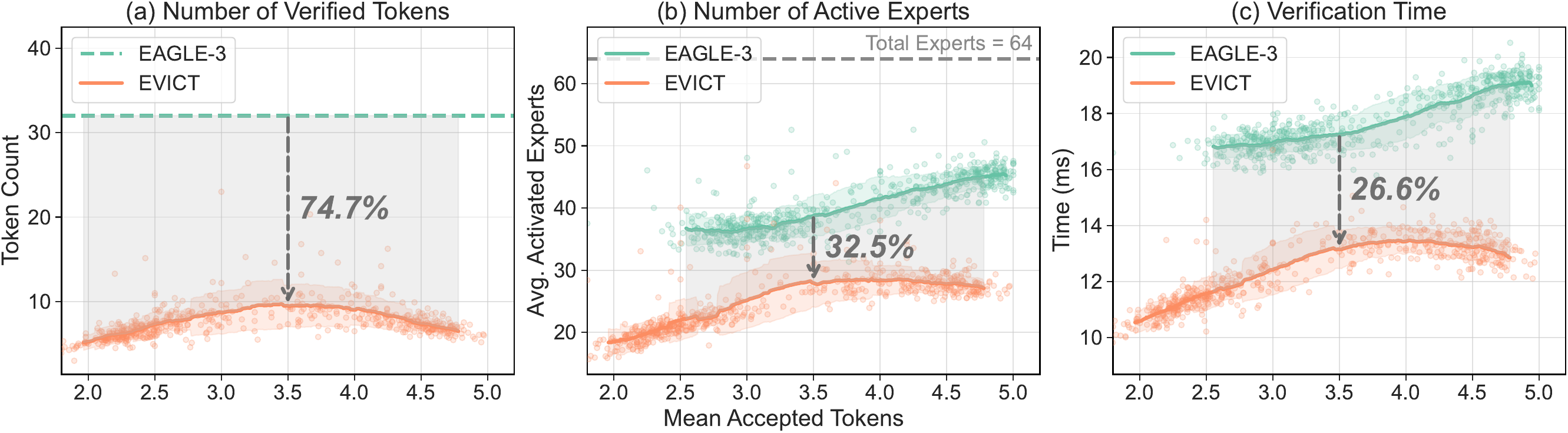}
    \caption{\textbf{Adaptive verification significantly reduces MoE verification cost.}
    The x-axis denotes the mean accepted tokens (\textsc{Mat}). By dynamically truncating low-value draft tokens across diverse contexts (a), EVICT substantially decreases the number of activated experts (b), which directly translates to lower target-model verification latency (c) compared to EAGLE-3's fixed-size trees.}
    \label{fig:three_efficiency_graphs}
\end{figure*}

\paragraph{Baselines and Implementation.} We compare EVICT with autoregressive decoding and three acceleration baselines, as detailed in Appendix~\ref{app:baselines}. All main experiments use a single-request setting (\texttt{batch\_size=1}) under the same SGLang framework~\citep{zheng2024sglang}; for EAGLE-family methods and EVICT, we use draft-model weights from SpecBundle~\citep{li2026specforge}. We report mean accepted tokens (\textsc{Mat}) and decoding speed (token/s), with further implementation details in Appendix~\ref{app:impl_details}. For high-concurrency scenario, Appendix~\ref{app:batch_results} reports larger-batch experiments (\texttt{batch\_size=4,8,16,32,64}) under a common vLLM setup with both EAGLE-3 and DFlash~\citep{chen2026dflash} drafters.

\begin{table}[t]
\centering
\setlength{\tabcolsep}{4pt}
\scalebox{0.85}{
\begin{tabular}{l|cc|cc}
\toprule
\multirow{2}{*}{Benchmark} & \multicolumn{2}{c|}{Temperature = 0} & \multicolumn{2}{c}{Temperature = 1} \\
& EAGLE-3 & EVICT & EAGLE-3 & EVICT \\
\midrule
Alpaca & 3.19 & 2.53 & 3.03 & 2.42 \\
GSM8K & 4.64 & 4.24 & 4.58 & 4.14 \\
HumanEval & 4.57 & 4.24 & 4.50 & 4.10 \\
MT-Bench & 3.46 & 2.79 & 3.24 & 2.64 \\
QA & 3.07 & 2.41 & 2.90 & 2.27 \\
CNN/DM & 2.95 & 2.33 & 2.84 & 2.29 \\
\midrule
\multirow{2}{*}{Average} & 3.65 & 3.09 & 3.52 & 2.98 \\
& (100\%) & (84.7\%) & (100\%) & (84.7\%) \\
\bottomrule
\end{tabular}
}
\caption{Mean accepted token (\textsc{Mat}) of Qwen3-30B-A3B on EVICT and EAGLE-3 across various benchmarks and temperatures. Values in parentheses indicate the ratio compared to EAGLE-3.}
\label{tab:mat_qwen_30b}
\end{table}

\subsection{Decoding Speed}
\label{sec:main_decoding_speed_result}

Table~\ref{tab:speed_qwen_30b} reports the main decoding-speed results on Qwen3-30B-A3B, while the full results on the other two MoE backbones are provided in Appendix~\ref{app:full_speed_results}. Overall, EVICT delivers substantial decoding acceleration, achieving speedups between $1.08\times$ and $2.35\times$ over standard autoregressive generation. These results reveal two primary advantages: \textbf{(I) Consistent superiority over the state of the art.} Across all evaluated settings, EVICT strictly outperforms the SOTA baseline EAGLE-3, as well as Lookahead and DDD, yielding a $1.21\times$ speedup over EAGLE-3 on average. The degradation of Lookahead suggests that MoE speculative decoding requires sufficiently accurate drafts to offset verification overhead, while the gap over DDD shows that confidence-based dynamic drafting alone is insufficient without a MoE-cost-aware strategy. \textbf{(II) Pronounced gains on math and coding.} On GSM8K and HumanEval, model outputs tend to be highly structured and predictable, leading to an inherently higher \textsc{Mat} (Table~\ref{tab:mat_qwen_30b}). EVICT consistently excels in these settings, maintaining high relative speedups over EAGLE-3 across temperatures, ranging from $1.23\times$ to $1.27\times$ on Qwen3-30B-A3B. These improvements confirm that our utility-guided tree truncation strategy effectively preserves high-value draft tokens while curtailing unnecessary MoE verification overhead. EVICT also remains beneficial on a dense target, with full results in Appendix~\ref{app:dense_results}.

\subsection{Trading \textsc{Mat} for Verification Overhead}
\label{sec:main_mat_result}

As demonstrated in Table~\ref{tab:mat_qwen_30b} and Appendix~\ref{app:full_mat_results}, EVICT generally yields a lower \textsc{Mat} compared to EAGLE-3. This reduction is particularly pronounced on benchmarks where EAGLE-3's acceptance rate is already relatively low (e.g., QA and CNN/DM). However, rather than a drawback, this highlights a key advantage of our approach: instead of relying on aggressively increasing draft acceptance, it adaptively selects a verification budget that better matches the benefit-to-cost ratio of each draft tree. By intentionally trading a marginal drop in \textsc{Mat} for a significant reduction in verification overhead, EVICT achieves higher overall decoding speed even in harder, low-acceptance regimes. To better understand this mechanism, we examine the intermediate statistics during decoding in Figure~\ref{fig:three_efficiency_graphs}.

\paragraph{Adaptive Token Verification.} Fig.~\ref{fig:three_efficiency_graphs} (a) compares the number of verified tokens at each step as a function of \textsc{Mat}. A striking pattern emerges: EVICT verifies 74.7\% fewer tokens than EAGLE-3 across the entire \textsc{Mat} range on average, with the strongest reduction appearing at both the low- and high-\textsc{Mat} extremes. When \textsc{Mat} is low (difficult contexts), additional draft tokens provide only marginal gains in expected acceptance but incur substantial verification overhead; EVICT discards these marginal tokens instead of verifying them. Conversely, when \textsc{Mat} is high (easy contexts), a small number of tokens is already sufficient to achieve a high acceptance length, rendering further verification of tail tokens unnecessary. Thus, EVICT dynamically adapts the verification budget to context difficulty, rather than applying a fixed policy.

\paragraph{Cost Reduction via Expert Sparsity.} Fig.~\ref{fig:three_efficiency_graphs} (b) and (c) illustrate the direct causal consequence of this adaptive truncation. Because EVICT verifies fewer tokens per step, the target MoE model needs to activate fewer unique experts. As shown in Fig.~\ref{fig:three_efficiency_graphs} (b), this substantially reduces the number of active experts by 32.5\% on average, compared with EAGLE-3. In MoE models, verification cost is dominated not only by token count but also by the diversity of experts loaded. Shrinking the union of activated experts across the draft tree alleviates memory bandwidth pressure during the forward pass. This translates directly into 26.6\% lower verification latency on average, as shown in Fig.~\ref{fig:three_efficiency_graphs} (c). This confirms that our advantage comes from selectively verifying only the tokens whose expected benefit justifies their cost.

\subsection{Ablation Study}

\subsubsection{Ablating Framework Overhead}

\begin{figure}[t]
    \centering
    \includegraphics[width=\columnwidth]{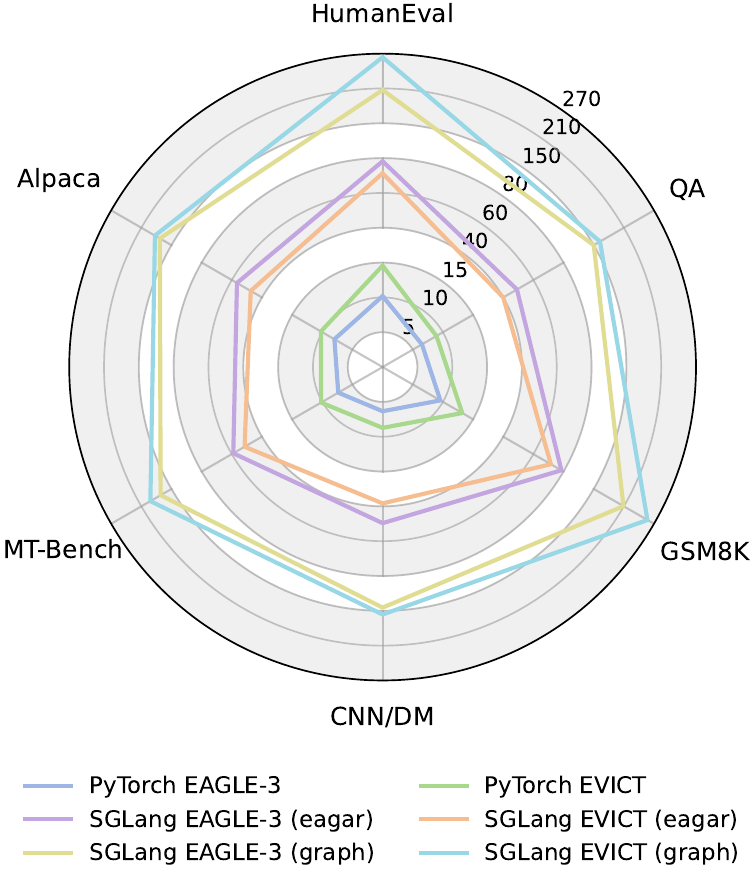}
    \caption{Decoding speed (token/s) of EVICT and EAGLE-3 for Qwen3-30B-A3B under different inference frameworks}
    \label{fig:radar}
\end{figure}

To fully realize the theoretical efficiency gains of our adaptive truncation strategy, it is critical to address the associated system-level overheads. Figure~\ref{fig:radar} presents a comprehensive ablation of decoding speed across six datasets, comparing our method against EAGLE-3 under three different implementation regimes: native PyTorch, SGLang eager execution, and SGLang with CUDA graphs.

\paragraph{The Necessity of System-Level Optimization.}
As illustrated in Figure~\ref{fig:radar}, relying solely on native PyTorch execution yields extremely low throughput for both methods, owing to inefficient compute kernels. While migrating to SGLang's eager mode provides a substantial baseline improvement, an interesting phenomenon emerges: EVICT slightly underperforms EAGLE-3 in this regime. This performance inversion occurs because our dynamic truncation strategy introduces additional runtime operations such as calculating the expected accepted length and finding the utility-maximizing prefix (Section~\ref{sec:choose_prefix}). In an eager execution environment, these extra CPU-bound calculations become a new bottleneck. Although our approach significantly reduces the GPU verification workload, the resulting latency reduction is entirely offset by the introduced CPU overhead.

\paragraph{Unleashing Performance via Graph Fusion.}
The situation reverses dramatically when CUDA graphs are enabled. As detailed in Section~\ref{sec:sgl_integrate}, we integrate our adaptive logic directly into SGLang by capturing target graphs of various lengths and fusing the tree truncation operations into the draft graph. This co-design of algorithm and system infrastructure completely eliminates the CPU overhead associated with dynamic tree pruning. Consequently, the GPU is unblocked to fully exploit the reduced verification cost. As shown in Figure~\ref{fig:radar}, our graph-fused implementation strictly outperforms EAGLE-3's graph implementation across all benchmarks, demonstrating that our adaptive verification algorithm translates into state-of-the-art end-to-end acceleration when properly supported by modern inference frameworks.

\begin{figure}[t]
    \centering
    \includegraphics[width=\columnwidth]{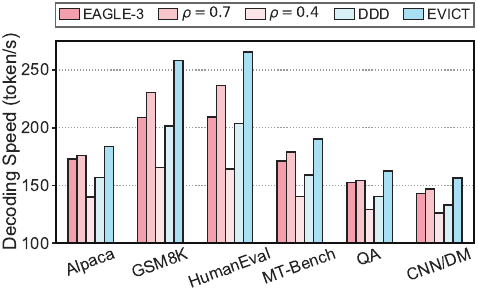}
    \caption{Decoding speed (token/s) of EVICT, DDD, EAGLE-3, and score-coverage variants with $\rho \in \{0.7, 0.4\}$ on Qwen3-30B-A3B.}
    \label{fig:ablate_method}
\end{figure}

\subsubsection{Ablating Cost-Aware Selection}

We ablate the cost-aware component of EVICT by removing the profiled verification cost $C(k)$ from the selection objective.
Specifically, we truncate the draft tree solely according to draft-score coverage: given $K$ total draft nodes sorted by cumulative score and $S_k=\sum_{i=1}^{k}\mathrm{Score}(v_i)$, we select the smallest prefix satisfying $S_k/S_K \ge \rho$.
Under this formulation, EAGLE-3 corresponds to the special case $\rho=1$, where all draft nodes under the fixed budget are verified.

As shown in Figure~\ref{fig:ablate_method}, both the score-coverage variants and DDD are cost-agnostic strategies that rely only on drafter-side confidence, and they consistently underperform EVICT.
Moreover, their performance depends on manually chosen thresholds such as $\rho$: aggressive truncation may discard useful draft nodes, while conservative truncation still retains costly low-value nodes.
In contrast, EVICT is hyperparameter-free for tree truncation and automatically selects the expected optimal cutoff by balancing estimated draft benefit with profiled MoE verification cost.
These results confirm that adaptive truncation alone is insufficient; explicitly modeling target-side verification cost is crucial for efficient MoE speculative decoding.

\section{Related Work}

\subsection{Mixture-of-Experts}

Mixture-of-Experts (MoE) improves the capacity-efficiency trade-off by activating only a small subset of parameters for each token. Following the sparse-gated formulation of \citep{shazeer2017outrageously}, subsequent work including GShard \citep{lepikhin2020gshard}, Switch Transformer \citep{fedus2022switch}, and GLaM \citep{du2022glam} established MoE as a scalable paradigm for large language models. Recent frontier-scale and highly sparse MoE models \citep{liu2024deepseek, zeng2026glm, team2025kimi, cao2026qwen3} further amplify systems pressure at inference time. In particular, MoE introduces nontrivial overheads in operator execution, expert-parallel communication, and load balancing, motivating specialized optimizations \citep{gale2023megablocks, mao2025uccl, han2025grace}. These characteristics make inference efficiency a central bottleneck in modern MoE systems \citep{liu2026survey} and directly motivate acceleration techniques such as speculative decoding.

\subsection{Tree-based Speculative Decoding}

Tree-based speculative decoding extends early chain-based drafting by verifying multiple candidates in parallel. Starting from SpecInfer \citep{miao2024specinfer}, subsequent work mainly improves draft-tree construction, including context-aware dynamic trees in EAGLE-2 \citep{li2024eagle} and adaptive tree optimization in OPT-Tree \citep{wang2025opt} and TALON \citep{liu2026talon}, to improve acceptance efficiency or reduce drafting cost. However, such methods primarily optimize drafting, while in large-scale MoE inference verification can become the dominant bottleneck, and highly dynamic strategies are often difficult to integrate with standard high-performance serving kernels. ECHO \citep{hu2026echo} addresses this issue in high-concurrency compute-bound settings, whereas our method instead targets the orthogonal regime of low-latency MoE speculative decoding.

\subsection{Speculative Decoding for MoE}

In memory-constrained offloading scenarios, SpecMoEOff~\cite{wang2025accelerating} and SP-MoE~\citep{wang2025sp} utilize speculative decoding to amortize CPU--GPU transfer and expert loading latency. For online serving, MoESD~\citep{huang2025moesd} shows that MoE-specific verification costs can dominate end-to-end efficiency, motivating MoE-aware optimization; however, it primarily analyzes when conventional speculative decoding benefits sparse MoE models rather than proposing a drop-in decoding algorithm. Utility-based adaptive speculation~\citep{saxena2025utility} mitigates this issue only through coarse-grained control of conventional chain-based speculation. MoE-Spec~\citep{mcdanel2026moe} reduces verification overhead via expert budgeting, but introduces an explicit quality-efficiency trade-off by truncating the active expert set. In contrast, our method leverages fine-grained drafter information and remains fully lossless without modifying the target model's verification semantics.
\section{Conclusion}

We introduce EVICT, a training-free and lossless MoE-cost-aware adaptive verification method that requires no manually tuned hyperparameters. EVICT selects the utility-maximizing tree prefix by balancing estimated accepted length against profiled average verification cost. Experiments on various MoE backbones and benchmarks show that EVICT consistently improves decoding speed over strong baselines, achieving an average 1.21$\times$ speedup over EAGLE-3. Additional results validate its behavior across batch sizes, drafting backbones, and a dense target model.

\section*{Limitations}

Although EVICT demonstrates strong acceleration for MoE speculative decoding, several limitations remain and motivate further investigation:

\paragraph{Robustness of Estimation.}
EVICT depends on estimating both the average verification cost of a tree prefix and its expected acceptance benefit. Our calibration and cost-stability results support these estimators across low- and high-acceptance workloads, but runtime conditions and domain shift may still affect their accuracy. A promising direction is to develop more robust and adaptive estimation strategies.

\paragraph{Effectiveness at Large Batch Sizes.}
EVICT's benefit at larger batch sizes is workload dependent. On low-\textsc{Mat} QA, its gains persist through batch size 64, whereas on high-\textsc{Mat} HumanEval most drafted tokens are already useful and EVICT remains broadly comparable to the untruncated backbones. This reduced headroom in high-acceptance or compute-bound regimes limits the benefit available from adaptive verification.

\bibliography{custom}

\newpage
\appendix

\section{Roofline Regimes}
\label{app:regime_analysis}

For a dense model, the roofline analysis yields two regimes:
\begin{enumerate}
    \item \textbf{Memory-bound.} Verifying multiple candidates reuses the same model weights, increasing arithmetic intensity approximately with number of verified tokens. Consequently, a multi-token verification pass can have latency close to that of a one-token pass. If $\ell$ tokens are accepted, the target-side speedup can approach $\ell$.
    \item \textbf{Compute-bound.} Once verification crosses the roofline ridge point, performance becomes limited by the hardware's peak compute throughput. Consequently, a multi-token verification pass takes longer than a one-token forward pass. If its latency is $p$ times that of a one-token pass, the target-side speedup decreases from the ideal $\ell$ to $\ell/p$.
\end{enumerate}

MoE models retain the compute-bound regime but further split the dense model's memory-bound regime according to expert saturation, expanding the above two regimes into three. Following MoESD~\citep{huang2025moesd}, under independent and approximately uniform routing, the expected number of unique experts activated by $t$ tokens is
\begin{equation}
    N_{\mathrm{uniq}}(t)
    = M\left[1-\left(1-\frac{K}{M}\right)^t\right],
\end{equation}
where $M$ is the total number of experts and $K$ is the number activated per token. This gives:
\begin{enumerate}
    \item \textbf{Expert-unsaturated, memory-bound.}\par When $t$ is small, $N_{\mathrm{uniq}}(t)\approx Kt$. computation and expert-weight traffic increase together. Arithmetic intensity improves little, and verification cost grows nearly linearly with the number of candidates.
    \item \textbf{Expert-saturated, memory-bound.}\par When $t$ is moderate, $N_{\mathrm{uniq}}(t)\approx M$, additional speculative tokens increasingly reuse already activated experts. Arithmetic intensity increases without a comparable increase in expert-weight traffic, making the marginal verification cost more dense-like and allowing the target-side speedup to again approach $\ell$.
    \item \textbf{Compute-bound.}\par At sufficiently high concurrency, verification latency scales approximately with the verified-token count, reducing the target-side speedup toward $\ell/p$.
\end{enumerate}

These regimes clarify the complementary scopes of existing approaches:
\begin{itemize}
    \item EVICT primarily targets expert-unsaturated, latency-oriented inference, where truncating low-utility nodes limits unnecessary expert-union growth.
    \item MoESD identifies the expert-saturated, moderate-batch regime in which MoE speculative decoding can be particularly favorable.
    \item ECHO~\citep{hu2026echo} targets high-concurrency, compute-bound serving through batch-level verification-budget allocation.
\end{itemize}

\section{Derivation of Expected Accepted Length}
\label{app:accepted_length_proof}

Let $A(\mathcal{T})$ denote the accepted length of a draft tree $\mathcal{T}$.
For each node $v \in \mathcal{T}$, define $I_v$ as the indicator that $v$ lies on the finally accepted path.
Then
\begin{equation}
    A(\mathcal{T}) = \sum_{v \in \mathcal{T}} I_v,
\end{equation}
and by linearity of expectation,
\begin{equation}
    \mathbb{E}[A(\mathcal{T})]
    =
    \sum_{v \in \mathcal{T}} \mathbb{P}(I_v = 1).
\end{equation}

Under the sequential tree verification rule described in Section~\ref{sec:tree_sd}, a node $v$ lies on the accepted path only if all nodes along the path from the root $x_{t+1}$ to $v$ are successively accepted.
Therefore,
\begin{equation}
    \mathbb{P}(I_v = 1)
    =
    \prod_{u \in \operatorname{Path}(x_{t+1}, v)}
    p_{t+1}^{\mathcal{T}}(u).
\end{equation}
Substituting this expression into the expectation above yields
\begin{equation}
    \mathbb{E}[A(\mathcal{T})]
    =
    \sum_{v \in \mathcal{T}}
    \prod_{u \in \operatorname{Path}(x_{t+1}, v)}
    p_{t+1}^{\mathcal{T}}(u),
\end{equation}
which proves Equation~\ref{eq:exact_accept_len}.

\section{Experimental Details}
\label{app:experimental_details}

\subsection{Estimator Validation}
\label{app:estimator_validation}

\begin{table*}[t]
\centering
\scriptsize
\resizebox{\textwidth}{!}{%
\begin{tabular}{llrrrrrrrrrr}
\toprule
Dataset & Metric & $[0,0.1)$ & $[0.1,0.2)$ & $[0.2,0.3)$ & $[0.3,0.4)$ & $[0.4,0.5)$ & $[0.5,0.6)$ & $[0.6,0.7)$ & $[0.7,0.8)$ & $[0.8,0.9)$ & $[0.9,1]$ \\
\midrule
\multirow{2}{*}{QA}
& Draft Confidence & 0.0715 & 0.1556 & 0.2501 & 0.3469 & 0.4446 & 0.5347 & 0.6489 & 0.7758 & 0.8810 & 0.9998 \\
& Target Acc. Rate & 0.0728 & 0.1832 & 0.2874 & 0.3796 & 0.4632 & 0.5537 & 0.6611 & 0.7848 & 0.8727 & 0.9817 \\
\midrule
\multirow{2}{*}{HumanEval}
& Draft Confidence & 0.0866 & 0.1633 & 0.2556 & 0.3486 & 0.4477 & 0.5347 & 0.6503 & 0.7773 & 0.8820 & 1.0000 \\
& Target Acc. Rate & 0.1540 & 0.1838 & 0.2961 & 0.3887 & 0.4445 & 0.5063 & 0.6577 & 0.7663 & 0.8861 & 0.9911 \\
\bottomrule
\end{tabular}}
\caption{Calibration of drafter confidence against empirical acceptance.}
\label{tab:calibration}
\end{table*}

\begin{table*}[t]
\centering
\small
\newcommand{\tstd}[1]{\,{\scriptsize $\pm #1$}}
\resizebox{\textwidth}{!}{%
\begin{tabular}{lrrrrrrrrr}
\toprule
Dataset & $k{=}1$ & $k{=}4$ & $k{=}8$ & $k{=}12$ & $k{=}16$ & $k{=}20$ & $k{=}24$ & $k{=}28$ & $k{=}32$ \\
\midrule
QA & $6.83$\tstd{0.15} & $8.02$\tstd{0.29} & $9.17$\tstd{0.34} & $9.43$\tstd{0.40} & $10.01$\tstd{0.38} & $10.77$\tstd{0.48} & $10.62$\tstd{0.49} & $10.36$\tstd{0.57} & $11.06$\tstd{0.56} \\
HumanEval & $7.04$\tstd{0.24} & $8.15$\tstd{0.28} & $9.75$\tstd{1.17} & $10.17$\tstd{0.52} & $10.70$\tstd{0.74} & $10.90$\tstd{0.52} & $11.16$\tstd{0.50} & $11.05$\tstd{0.68} & $11.40$\tstd{0.61} \\
\bottomrule
\end{tabular}}
\caption{Profiled speculative-step latency in milliseconds (mean $\pm$ standard deviation).}
\label{tab:cost_stability}
\end{table*}

\paragraph{Benefit Estimation.}
We group draft tokens into ten intervals by their draft confidence (conditional probability given by draft model) and compare mean confidence with real acceptance rate after target verification. As shown in Table~\ref{tab:calibration}, acceptance increases monotonically with confidence on both QA and HumanEval, supporting the use of drafter confidence to rank candidate benefit. The expected calibration errors (ECEs) are 0.0204 and 0.0107, respectively. ECE measures the gap between predicted confidence and empirical acceptance frequency, so these small values show that the proxy remains well calibrated in both low- and high-\textsc{Mat} regimes. The individual-outcome Brier scores are 0.1659 and 0.0731; the lower value on HumanEval is consistent with its more predictable high-\textsc{Mat} workload, whereas QA exhibits greater per-node uncertainty despite retaining good aggregate calibration.

\paragraph{Cost Estimation.}
We profile every verification length $k\in[1,32]$ on QA and HumanEval, using 32 measurements for each $k$. Table~\ref{tab:cost_stability} reports the single-token cost and every fourth length. Across all 32 profiled points, the raw cost curves have Pearson correlation 0.961, showing strong linear agreement in how latency grows with $k$, and Spearman correlation 0.924, showing that the relative ordering of verification lengths by cost is also well preserved. The MAE of 0.57 ms and MAPE of 5.84\% quantify the remaining absolute mismatch between workloads. Because uniformly scaling $C(k)$ does not change the maximizing $k$, the profile shape is more important for EVICT's decision than its absolute level. After mean normalization, the MAE is only 2.63\%, and the relative growth is nearly identical, with $C(32)/C(1)=1.620$ on QA and 1.619 on HumanEval. Together, these statistics indicate that offline profiling captures a stable cost shape despite workload-dependent routing variation, supporting its use for online prefix selection.

\subsection{Model Configurations}
\label{app:models}

We evaluate EVICT on three MoE backbones with diverse model scales and routing sparsity patterns, together with one dense target model.
Specifically, Qwen3-30B-A3B has 30B total parameters, 3B active parameters, 128 experts, and activates 8 experts per token~\citep{qwen-3};
Qwen3-235B-A22B has 235B total parameters, 22B active parameters, 128 experts, and activates 8 experts per token~\citep{qwen-3};
and Ling-flash-2.0 has 103B total parameters, 6.1B active parameters, 256 experts, and activates 8 experts per token~\citep{lingteam2025activationboostedscalinggeneral}.
We additionally use Hunyuan-7B-Instruct, a dense instruction-tuned model, to test whether EVICT's adaptive verification benefits generalize beyond sparse expert routing.

To ensure full reproducibility, Table~\ref{tab:model_weights} lists the exact Hugging Face checkpoints used in our experiments.
For a strictly fair comparison with EAGLE-3, all EAGLE-family methods, including EAGLE-3, DDD, and EVICT, use the same target-model checkpoints and the same EAGLE-3 draft-model checkpoints.
Therefore, any performance differences among these methods can be attributed to the decoding and verification strategy rather than differences in draft-model weights.

\begin{table*}[t]
\centering
\small
\tabcolsep=0.8em
\scalebox{0.88}{
\begin{tabular}{ll}
\toprule
Target Model Checkpoint & EAGLE-3 Draft Model Checkpoint \\
\midrule
\texttt{Qwen/Qwen3-30B-A3B-Instruct-2507}
& \texttt{lmsys/SGLang-EAGLE3-Qwen3-30B-A3B-Instruct-2507-SpecForge-Nex} \\
\texttt{inclusionAI/Ling-flash-2.0}
& \texttt{AQ-MedAI/Ling-Flash-2.0-eagle3} \\
\texttt{Qwen/Qwen3-235B-A22B-Instruct-2507}
& \texttt{lmsys/SGLang-EAGLE3-Qwen3-235B-A22B-Instruct-2507-SpecForge-Meituan} \\
\texttt{tencent/Hunyuan-7B-Instruct}
& \texttt{AngelSlim/Hunyuan-7B-Instruct\_eagle3} \\
\bottomrule
\end{tabular}
}
\caption{
Exact Hugging Face checkpoints for target models and their corresponding EAGLE-3 draft models.
}
\label{tab:model_weights}
\end{table*}

\subsection{Baselines}
\label{app:baselines}

\paragraph{Vanilla.}
Standard autoregressive decoding without speculative acceleration.

\paragraph{Lookahead.}
Lookahead~\citep{zhao2024lookahead} is a lossless multi-branch decoding framework that does not rely on an auxiliary draft model.
It retrieves candidate continuations from previous contexts with a trie structure and verifies multiple branches in parallel with the target model.

\paragraph{EAGLE-3.}
EAGLE-3~\citep{eagle3} is a state-of-the-art tree-based speculative decoding method.
It improves draft quality through direct token prediction and multi-layer feature fusion, then constructs a draft token tree for parallel target verification.

\paragraph{Dynamic Depth Decoding (DDD).}
Based on EAGLE-2, DDD~\citep{brown2024dynamic} extends tree drafting with confidence-based dynamic depth.
It adaptively stops tree expansion according to drafter confidence, but does not explicitly model target-side MoE verification cost.

\subsection{Implementation Details}
\label{app:impl_details}

All experiments are conducted using the SGLang inference engine~\citep{zheng2024sglang} on NVIDIA A100 SXM4 80GB GPUs.
For models that do not fit on a single GPU, we enable tensor parallelism.
Specifically, Ling-flash-2.0 is evaluated with \texttt{TP=4}, and Qwen3-235B-A22B is evaluated with \texttt{TP=8}.

For EAGLE-3, DDD, and EVICT, we adopt the same base tree configuration with \texttt{steps=4}, \texttt{topk=8}, and \texttt{draft\_tokens=32}.
This configuration provides a balanced trade-off among tree depth, tree width, and verification overhead.
Lookahead and EAGLE-3 are already integrated in SGLang, while DDD is not officially supported.
We therefore reimplement DDD by applying its dynamic-depth policy on top of the same EAGLE-3 drafting pipeline in SGLang, using the same draft weights and tree configuration as EAGLE-3 and EVICT for fair comparison.

\subsection{Larger-Batch Evaluation}
\label{app:batch_results}

We evaluate EVICT beyond the single-request setting on QA and HumanEval with batch sizes from 4 to 64. All methods use the same vLLM setup and aggregate decoding throughput metric. EVICT's variable verification lengths are supported through piecewise CUDA Graph execution. We apply the same EVICT selection rule to both EAGLE-3 and DFlash drafting backbones. Table~\ref{tab:batch_results} reports the complete results.

On low-\textsc{Mat} QA, EVICT consistently improves both backbones across all batch sizes, with gains reaching 34.7\% over EAGLE-3 and 39.1\% over DFlash at BS=64. On high-\textsc{Mat} HumanEval, where most draft tokens are already useful, EVICT has less headroom: it largely preserves EAGLE-3's strong acceleration and improves DFlash at BS=8--64. Overall, EVICT remains effective beyond batch size 1, particularly in low-\textsc{Mat} regimes.

\begin{table}[t]
\centering
\scriptsize
\resizebox{\columnwidth}{!}{%
\begin{tabular}{lrrrrr}
\toprule
\multicolumn{6}{c}{\textbf{QA (low \textsc{Mat})}} \\
\cmidrule(lr){1-6}
Method & BS=4 & BS=8 & BS=16 & BS=32 & BS=64 \\
\midrule
Vanilla AR & 454 & 754 & 1,280 & 2,126 & 3,519 \\
\midrule
EAGLE-3 & 569 & 984 & 1,530 & 2,228 & 2,935 \\
EAGLE-3 + EVICT & \textbf{598} & \textbf{1,026} & \textbf{1,724} & \textbf{2,763} & \textbf{3,953} \\
\midrule
DFlash & 603 & 1,015 & 1,610 & 2,245 & 2,852 \\
DFlash + EVICT & \textbf{662} & \textbf{1,110} & \textbf{1,805} & \textbf{2,823} & \textbf{3,968} \\
\midrule
\multicolumn{6}{c}{\textbf{HumanEval (high \textsc{Mat})}} \\
\cmidrule(lr){1-6}
Method & BS=4 & BS=8 & BS=16 & BS=32 & BS=64 \\
\midrule
Vanilla AR & 435 & 742 & 1,249 & 2,109 & 3,418 \\
\midrule
EAGLE-3 & 1,387 & \textbf{2,583} & \textbf{4,299} & \textbf{6,223} & \textbf{8,298} \\
EAGLE-3 + EVICT & \textbf{1,416} & 2,420 & 3,994 & 6,020 & 8,215 \\
\midrule
DFlash & 1,398 & 2,044 & 3,572 & 5,252 & 6,354 \\
DFlash + EVICT & 1,226 & \textbf{2,098} & \textbf{3,638} & \textbf{5,568} & \textbf{6,448} \\
\bottomrule
\end{tabular}}
\caption{Aggregate decoding throughput (token/s) across batch sizes in the common vLLM setup. EVICT is applied to both EAGLE-3 and DFlash drafting backbones.}
\label{tab:batch_results}
\end{table}

\subsection{More Results on Decoding Speed}
\label{app:full_speed_results}

Tables~\ref{tab:speed_ling} and~\ref{tab:speed_qwen_235b} report additional decoding-speed results on Ling-flash-2.0 and Qwen3-235B-A22B.
Consistent with the main results in Section~\ref{sec:main_decoding_speed_result}, EVICT achieves the best decoding speed across all benchmarks and temperatures.
Compared with EAGLE-3, EVICT further improves average decoding speed by $1.22\times$ on Ling-flash-2.0 and $1.25\times$ on Qwen3-235B-A22B.
Lookahead remains below vanilla decoding, and DDD consistently lags behind EVICT, further confirming that accurate drafting and cost-aware verification are both crucial for efficient MoE speculative decoding.

\begin{table*}[t]
\centering
\tabcolsep=0.6em
\scalebox{0.83}{
\begin{tabular}{l|*{6}{>{\centering\arraybackslash}p{4.55em}}|c}
\toprule
Method & Alpaca & GSM8K & HumanEval & QA & MT-Bench & CNN/DM & Average \\
\midrule
\multicolumn{8}{c}{Temperature = 0} \\
\midrule
Vanilla & 179.80 & 174.90 & 174.51 & 175.32 & 174.57 & 174.37 & 175.58 (1.00$\times$) \\
Lookahead & 118.59 & 138.02 & 130.89 & 121.47 & 133.59 & 104.45 & 124.50 (0.71$\times$) \\
\midrule
EAGLE-3 & 225.62 & 255.58 & 254.55 & 198.58 & 221.97 & 193.60 & 224.98 (1.28$\times$) \\
\midrule
+DDD & 217.71 & 253.64 & 253.06 & 187.23 & 212.33 & 180.66 & 217.44 (1.24$\times$) \\
\textbf{+EVICT} & \textbf{267.44} & \textbf{348.52} & \textbf{351.07} & \textbf{225.28} & \textbf{270.96} & \textbf{216.51} & \textbf{279.96 (1.59$\times$)} \\
\midrule
\multicolumn{8}{c}{Temperature = 1} \\
\midrule
Vanilla & 174.50 & 170.31 & 169.86 & 170.78 & 170.29 & 170.29 & 171.01 (1.00$\times$) \\
Lookahead & 119.35 & 135.04 & 128.99 & 117.18 & 129.01 & 102.95 & 122.09 (0.71$\times$) \\
\midrule
EAGLE-3 & 216.64 & 250.50 & 249.55 & 188.61 & 213.12 & 185.41 & 217.30 (1.27$\times$) \\
\midrule
+DDD & 209.16 & 249.89 & 244.41 & 180.36 & 203.17 & 174.18 & 210.19 (1.23$\times$) \\
\textbf{+EVICT} & \textbf{249.98} & \textbf{336.81} & \textbf{335.78} & \textbf{208.96} & \textbf{254.83} & \textbf{208.19} & \textbf{265.76 (1.55$\times$)} \\
\bottomrule
\end{tabular}
}
\caption{Decoding speed (token/s) on Ling-flash-2.0 across benchmarks and temperatures. Values in parentheses indicate speedup over vanilla autoregressive decoding.}
\label{tab:speed_ling}
\end{table*}
\begin{table*}[t]
\centering
\tabcolsep=0.6em
\scalebox{0.83}{
\begin{tabular}{l|*{6}{>{\centering\arraybackslash}p{4.55em}}|c}
\toprule
Method & Alpaca & GSM8K & HumanEval & QA & MT-Bench & CNN/DM & Average \\
\midrule
\multicolumn{8}{c}{Temperature = 0} \\
\midrule
Vanilla & 82.69 & 79.61 & 80.61 & 84.11 & 80.68 & 83.64 & 81.89 (1.00$\times$) \\
Lookahead & 59.53 & 65.50 & 63.32 & 57.01 & 64.73 & 50.69 & 60.13 (0.73$\times$) \\
\midrule
EAGLE-3 & 120.22 & 146.38 & 144.34 & 96.31 & 112.68 & 89.49 & 118.24 (1.44$\times$) \\
\midrule
+DDD & 109.68 & 137.71 & 137.64 & 92.17 & 107.17 & 89.14 & 112.25 (1.37$\times$) \\
\textbf{+EVICT} & \textbf{143.73} & \textbf{187.43} & \textbf{188.12} & \textbf{117.17} & \textbf{137.81} & \textbf{111.89} & \textbf{147.69 (1.80$\times$)} \\
\midrule
\multicolumn{8}{c}{Temperature = 1} \\
\midrule
Vanilla & 81.99 & 79.01 & 76.73 & 83.65 & 80.99 & 82.56 & 80.82 (1.00$\times$) \\
Lookahead & 59.52 & 64.60 & 62.96 & 55.67 & 63.50 & 50.46 & 59.45 (0.74$\times$) \\
\midrule
EAGLE-3 & 119.65 & 143.43 & 141.31 & 93.15 & 108.28 & 88.14 & 115.66 (1.43$\times$) \\
\midrule
+DDD & 108.91 & 135.65 & 135.70 & 90.77 & 105.36 & 86.97 & 110.56 (1.37$\times$) \\
\textbf{+EVICT} & \textbf{140.46} & \textbf{184.44} & \textbf{183.21} & \textbf{112.48} & \textbf{134.18} & \textbf{108.98} & \textbf{143.96 (1.78$\times$)} \\
\bottomrule
\end{tabular}
}
\caption{Decoding speed (token/s) on Qwen3-235B-A22B across benchmarks and temperatures. Values in parentheses indicate speedup over vanilla autoregressive decoding.}
\label{tab:speed_qwen_235b}
\end{table*}

\subsection{Dense-Model Comparison}
\label{app:dense_results}

We additionally evaluate EVICT on the dense Hunyuan-7B-Instruct target model. Table~\ref{tab:dense_results} shows that EVICT improves EAGLE-3 on every benchmark by 2.6\%--11.3\%, demonstrating that adaptive verification also generalizes to dense models. Since the full EAGLE-3 tree $k=K$ (number of verified tokens equals to size of whole tree) remains a feasible candidate, EVICT can retain the original verification budget when truncation is not beneficial; empirically, it does not regress on any workload. The gains are smaller than those observed on MoE targets, consistent with EVICT's primary motivation of reducing the additional expert-union cost in sparse MoE verification.

\begin{table*}[t]
\centering
\tabcolsep=0.6em
\scalebox{0.83}{
\begin{tabular}{l|*{6}{>{\centering\arraybackslash}p{4.55em}}|c}
\toprule
Method & Alpaca & GSM8K & HumanEval & QA & MT-Bench & CNN/DM & Average \\
\midrule
Vanilla AR & 138.79 & 140.69 & 139.94 & 141.33 & 140.98 & 142.05 & 140.63 (1.00$\times$) \\
\midrule
EAGLE-3 & 139.92 & 231.45 & 245.45 & 143.27 & 194.84 & 139.55 & 182.41 (1.30$\times$) \\
\textbf{+EVICT} & \textbf{154.19} & \textbf{256.94} & \textbf{264.84} & \textbf{156.86} & \textbf{200.17} & \textbf{151.87} & \textbf{197.48 (1.40$\times$)} \\
\bottomrule
\end{tabular}
}
\caption{Decoding speed (token/s) on the dense Hunyuan-7B target at Temperature $=0$. Values in parentheses indicate speedup over vanilla autoregressive decoding. EVICT improves EAGLE-3 on all six benchmarks.}
\label{tab:dense_results}
\end{table*}

\subsection{More Results on \textsc{Mat}}
\label{app:full_mat_results}

Tables~\ref{tab:mat_ling} and~\ref{tab:mat_qwen3_235b} report additional \textsc{Mat} results on Ling-flash-2.0 and Qwen3-235B-A22B.
EVICT consistently retains most of EAGLE-3's accepted length while verifying fewer draft nodes, matching the trend observed in Section~\ref{sec:main_mat_result}.
Together with the decoding-speed results, these results confirm that EVICT improves end-to-end efficiency by trading a modest reduction in \textsc{Mat} for a larger reduction in MoE verification overhead.

\begin{table}[t]
\centering
\setlength{\tabcolsep}{4pt}
\scalebox{0.85}{
\begin{tabular}{l|cc|cc}
\toprule
\multirow{2}{*}{Benchmark} & \multicolumn{2}{c|}{Temperature = 0} & \multicolumn{2}{c}{Temperature = 1} \\
& EAGLE-3 & EVICT & EAGLE-3 & EVICT \\
\midrule
Alpaca & 3.72 & 3.20 & 3.56 & 3.05 \\
GSM8K & 4.71 & 4.48 & 4.71 & 4.37 \\
HumanEval & 4.75 & 4.53 & 4.69 & 4.39 \\
MT-Bench & 3.76 & 3.27 & 3.60 & 3.14 \\
QA & 3.39 & 2.95 & 3.27 & 2.80 \\
CNN/DM & 3.31 & 2.85 & 3.21 & 2.78 \\
\midrule
\multirow{2}{*}{Average} & 3.94 & 3.55 & 3.84 & 3.42 \\
& (100\%) & (90.1\%) & (100\%) & (89.1\%) \\
\bottomrule
\end{tabular}
}
\caption{Mean accepted tokens (\textsc{Mat}) on Ling-flash-2.0 across benchmarks and temperatures. Values in parentheses indicate ratios relative to EAGLE-3.}
\label{tab:mat_ling}
\end{table}
\begin{table}[t]
\centering
\setlength{\tabcolsep}{4pt}
\scalebox{0.85}{
\begin{tabular}{l|cc|cc}
\toprule
\multirow{2}{*}{Benchmark} & \multicolumn{2}{c|}{Temperature = 0} & \multicolumn{2}{c}{Temperature = 1} \\
& EAGLE-3 & EVICT & EAGLE-3 & EVICT \\
\midrule
Alpaca & 3.18 & 2.61 & 3.12 & 2.54 \\
GSM8K & 4.58 & 3.99 & 4.43 & 4.00 \\
HumanEval & 4.56 & 4.11 & 4.53 & 4.03 \\
MT-Bench & 3.39 & 2.81 & 3.28 & 2.71 \\
QA & 2.90 & 2.38 & 2.85 & 2.34 \\
CNN/DM & 2.82 & 2.34 & 2.77 & 2.30 \\
\midrule
\multirow{2}{*}{Average} & 3.57 & 3.04 & 3.50 & 2.99 \\
& (100\%) & (85.2\%) & (100\%) & (85.4\%) \\
\bottomrule
\end{tabular}
}
\caption{Mean accepted tokens (\textsc{Mat}) on Qwen3-235B-A22B across benchmarks and temperatures. Values in parentheses indicate ratios relative to EAGLE-3.}
\label{tab:mat_qwen3_235b}
\end{table}

\section{LLM Usage}

We used LLM-based assistants during the preparation of this work for language polishing, wording suggestions, assistance in understanding related papers, and suggestions on possible code modifications. The assistants were not used to generate the research idea, design the algorithm, conduct experiments, implement the method, analyze results, or make scientific claims. All algorithmic decisions, code implementation, experimental analysis, and final writing decisions were made and verified by the authors.

LLM-based assistants are not listed as authors, as they do not meet the authorship criteria. The authors take full responsibility for all content in this paper.

\end{document}